\documentclass[conference]{IEEEtran}
\IEEEoverridecommandlockouts

\usepackage{cite}
\usepackage{amsmath,amssymb,amsfonts}
\usepackage{algorithm}
\usepackage{algorithmic}
\usepackage{graphicx}
\usepackage{textcomp}
\usepackage{xcolor}
\def\BibTeX{{\rm B\kern-.05em{\sc i\kern-.025em b}\kern-.08em
    T\kern-.1667em\lower.7ex\hbox{E}\kern-.125emX}}

\newtheorem{lemma}{Lemma}
\newtheorem{proposition}{Proposition}
\newtheorem{corollary}{Corollary}

\newtheorem{assumption}{Assumption}

\begin{document}

\title{Algorithm Design for Continual Learning  in IoT Networks

\thanks{This work has been accepted to be presented at IEEE ICASSP'25.}

\thanks{This work is supported in part by the Ministry of Education, Singapore, under its Academic Research Fund Tier 2 Grant with Award no. MOE-T2EP20121-0001; in part by SUTD Kickstarter Initiative (SKI) Grant with no. SKI 2021\_04\_07; and in part by the Joint SMU-SUTD Grant with no. 22-LKCSB-SMU-053.}
 }

\author{\IEEEauthorblockN{Shugang Hao, Lingjie Duan}
\IEEEauthorblockA{\textit{Pillar of Engineering Systems and Design, Singapore University of Technology and Design} \\
shugang\_hao@sutd.edu.sg, lingjie\_duan@sutd.edu.sg}
}

\maketitle

\begin{abstract}

Continual learning (CL) is a new online learning technique over sequentially generated streaming data from different tasks, aiming to maintain a small forgetting loss on previously-learned tasks. Existing work focuses on reducing the forgetting loss under a given task sequence. However, if similar tasks continuously appear to the end time, the forgetting loss is still huge on prior distinct tasks. In practical IoT networks, an autonomous vehicle to sample data and learn different tasks can route and alter the order of task pattern at increased travelling cost. To our best knowledge, we are the first to study how to opportunistically route the testing object and alter the task sequence in CL. We formulate a new optimization problem and prove it NP-hard. We propose a polynomial-time algorithm to achieve approximation ratios of $\frac{3}{2}$ for underparameterized case and $\frac{3}{2} + r^{1-T}$ for overparameterized case, respectively.
Simulation results verify our algorithm’s close-to-optimum performance.

\end{abstract}


\section{Introduction}

Continual learning (CL) is a new online learning technique over sequentially generated streaming data from different tasks and aims to maintain a small forgetting loss on previously-learned tasks (e.g., \cite{parisi2019continual}). A critical challenge in CL is catastrophic forgetting (e.g., \cite{wang2024comprehensive}), where the agent incurs a degraded performance on previous tasks after being trained on a new one.  To address this issue, many studies are proposed in the CL literature for new task learning, such as regularizing old tasks' weights (e.g., \cite{wang2023task}, \cite{zhao2024statistical}, \cite{kirkpatrick2017overcoming}), expanding the neural network (e.g., \cite{yoon2017lifelong}, \cite{zhang2024topology}, \cite{li2024theory}) and replaying old tasks' data  (e.g., \cite{lopez2017gradient}, \cite{shi2024unified}, \cite{nie2023bilateral}). 

However, the existing literature focuses on reducing the forgetting loss under a given data or task sequence. If similar tasks continuously appear to the end time, the forgetting loss is still huge on the prior distinct tasks (e.g., \cite{bell2022effect}). Few studies analyze the effect of task pattern on forgetting loss, which only consider preliminary cases with a limited number of tasks or some specific task patterns  (e.g., \cite{bell2022effect}, \cite{evron2022catastrophic}, \cite{lin2023theory}).

In practical IoT networks, an autonomous agent to sample data and learn tasks under different contexts can actually route and alter the order of data task pattern at increased travelling cost.
For example, a warehouse robot 
decides how to route items throughout the facility when handling online tasks like picking and packing for forgetting loss minimization (e.g., \cite{shaheen2022continual}). Further, Evron \textit{et al.} (2022) in \cite{evron2022catastrophic} presents a real-world example of autonomous vehicle training. An agent wants to learn a predictor for pedestrian detection in an autonomous vehicle, which is required to operate well in $T$ geographically distant  regions of different landscapes (e.g., city, forest, desert). He actively determines how to go through the $T$ regions for data sampling and training to maintain a good prediction performance on each region. 
 
Nonetheless, it is challenging for an agent to determine the order of data or task pattern in the IoT network in CL. Since tasks are geographically distant, it incurs a travelling cost when moving in the IoT network (e.g., warhouse robots or autonomous vehicles). Further, the training data is generated in an online manner, which cannot be known by the agent only after his routing decision. Later we prove that the optimization problem is NP-hard, which cannot be solved optimally in a polynomial time. It is thus required to propose a good approximation algorithm for a small ratio to the optimum.

To our best knowledge, we are the first to study \textit{how to opportunistically route the testing object and alter the task sequence in CL}. We formulate this as a new optimization problem and prove it NP-hard. We propose an algorithm to achieve approximation ratios of $\frac{3}{2}$ for the underparameterized case and $\frac{3}{2} + r^{1-T}$ for the overparameterized case in a polynomial time in task number $T$, respectively, where $r:=1-\frac{n}{m}$ is a parameter of feature number $m$ and sample number $n$ and $T$ is the task number.
Simulation results verify our algorithm’s close-to-optimum performance. 

\section{System Model and Problem Formulation}

First, we introduce our system model of continual learning in an IoT network. Then, we formulate our optimization problem and present an essential assumption for analysis.

\subsection{System Model of Continual Learning in the IoT Network}

We use the motivating example in \cite{evron2022catastrophic} to illustrate our system model for ease of understanding. We consider an agent who wants to learn a predictor for pedestrian detection in an autonomous vehicle, which is required to perform well in $T$ different geographical regions. Define $\boldsymbol{\tau} := (\tau_1, \cdots, \tau_T)$ as a sequential route among the $T$ regions or tasks. At each region $\tau_t$ for $t \in [T]:=\{1, \cdots, T\}$, he drives the vehicle to collect data $(\boldsymbol{x}_{\tau_t}, \boldsymbol{y}_{\tau_t})$ (e.g., by taking images of all the pedestrians in the region) for the predictor training in the region. 

We denote $\boldsymbol{x}_{\tau_t}$ as an $m \times n$ feature vector with $m$ features (e.g., image pixel) and $n$ samples. The agent only knows that each element $x_{\tau_t}^{(i, j)}$ of $\boldsymbol{x}_{\tau_t}$ is independent of each other for $i$$\in$$[m]$ and $j$$\in$$[n]$, which follows a normal distribution as $x_{\tau_t}^{(i, j)}\sim\mathcal{N}(0,1)$ (e.g., \cite{evron2022catastrophic}, \cite{lin2023theory}). We denote $\boldsymbol{y}_{\tau_t}$ as the $n \times 1$ output vector (e.g., score of pedestrian prediction). Following the CL and IoT literature (e.g., \cite{evron2022catastrophic}, \cite{lin2023theory}), we model that each output vector $\boldsymbol{y}_{\tau_t}$ is realized from a linear regression model:
\begin{align*}
    \boldsymbol{y}_{\tau_t} = \boldsymbol{x}_{\tau_t}^T\boldsymbol{w}_{\tau_t}^* + \boldsymbol{z}_{\tau_t},
\end{align*}
where $\boldsymbol{w}_{\tau_t}^*$ is the $m \times n$ vector of the ground-truth model parameters (e.g., feature weights) and $\boldsymbol{z}_{\tau_t}$ is the $n \times 1$ noise vector (e.g., image background). Each element $z_{\tau_t}^{(i)}$ of $\boldsymbol{z}_{\tau_t}$ is independent of each other for $i \in [n]$ and follows a Gaussian distribution as $z_{\tau_t}^{(i)}\sim\mathcal{N}(0,\sigma^2)$ (e.g., \cite{evron2022catastrophic}, \cite{lin2023theory}).  

The agent only knows the distribution of the noise and is uncertain of neither the ground-truth vector $\boldsymbol{w}_{\tau_t}^*$ nor the noise realization $\boldsymbol{z}_{\tau_t}$. In practice, an autonomous device may have low computational and memory resources, preventing it from storing data of old tasks (e.g., \cite{shaheen2022continual}). We then consider a memoryless setting to learn each predictor $\boldsymbol{w}_{\tau_t}$ by minimizing the training loss given the obtained data $(\boldsymbol{x}_{\tau_t}, \boldsymbol{y}_{\tau_t})$ as follows:
\begin{align}\label{e1}
    L_{\tau_t}(\boldsymbol{w}) = \frac{1}{n} ||\boldsymbol{x}_{\tau_t}^T \boldsymbol{w} - \boldsymbol{y}_{\tau_t} ||_2^2.
\end{align}

In Sections~\ref{S3} and \ref{S4}, we analyze both the underparameterized case (i.e., $m \leq n-2$) and overparameterized case (i.e., $m \geq n + 2$). The cases of $m \in \{n-1, n, n+1\}$ are undefined for each Inverse-Wishart distributed $(\boldsymbol{x}_t^T\boldsymbol{x}_t)^{-1}$, which is essential to obtain the solution $\boldsymbol{w}_{\tau_t}$ to \eqref{e1} (e.g., \cite{muirhead2009aspects}).

The agent needs to determine how to route in the region graph for continuous data sampling and training to maintain a small forgetting loss over all previously-visited regions and then actively alters the order of data task pattern at increased travelling cost. We denote $c_{i,j}$ as the travelling distance or cost between regions $i$ and $j$, which satisfies $c_{i,j} \leq c_{i,k} + c_{k,j}$ for any $i,j,k \in [T]$ and $i\ne j \ne k$.

As the agent may still receive tasks from previously-visited regions in the future, he expects a small generalized loss of his final predictor $\boldsymbol{w}_{\tau_T}$ in the last training region $\tau_T$ from all the ground-truth model parameters $\{\boldsymbol{w}_{\tau_t}^*\}_{t=1}^T$. We thus define his forgetting loss $F_T$ in all $T$ regions as the average sum of the squared $\ell_2$-norm distance as follows (e.g., \cite{lin2023theory}):
\begin{align}\label{e2}
    F_T(\boldsymbol{w}_{\tau_T}) = \frac{1}{T} \sum_{t=1}^{T} ||\boldsymbol{w}_{\tau_T} - \boldsymbol{w}_{\tau_t}^*||^2.
\end{align}
When routing in the IoT network, he incurs a  travelling cost $\sum_{t=1}^{T-1}c_{\tau_t,\tau_{t+1}}$. Thus, we define his overall loss $\pi(\boldsymbol{\tau})$ as the sum of the forgetting loss in \eqref{e2} and average travelling cost among the $T$ regions as follows:
\begin{align}\label{e4}
    \pi(\boldsymbol{\tau}) = \frac{1}{T} \sum_{t=1}^{T} ||\boldsymbol{w}_{\tau_T} - \boldsymbol{w}_{\tau_t}^*||^2+ \frac{1}{T} \sum_{t=1}^{T-1}c_{\tau_t,\tau_{t+1}}.
\end{align}
Note that the agent cannot control any online data input $(x_{\tau_t}, y_{\tau_t})$ but the travelling order $\boldsymbol{\tau}$.

\subsection{Problem Formulation}

In practice, each ground-truth parameter $\boldsymbol{w}_{\tau_i}^*$ is unknown to the agent (e.g., \cite{lin2023theory}). Thus, the dissimilarity of ground-truth parameters $||\boldsymbol{w}_{\tau_i}^*-\boldsymbol{w}_{\tau_j}^*||^2$ between regions $\tau_i$ and $\tau_j$ is unknown to the agent. Further,  the dissimilarity between each region $\tau_t$'s ground-truth parameter $\boldsymbol{w}_{\tau_t}^*$ and the agent's initial predictor $\boldsymbol{w}_{0}$ is unknown to the agent, either. Similar to \cite{lin2023theory}, we assume without loss of generality in the following.
\begin{assumption}\label{a1}
    The dissimilarity between ground-truth parameters $\boldsymbol{w}_{\tau_i}^*$ of region $\tau_i$ and $\boldsymbol{w}_{\tau_j}^*$ of region $\tau_j$ is upper-bounded as $||\boldsymbol{w}_{\tau_i}^*-\boldsymbol{w}_{\tau_j}^*||^2 \leq \Delta_{\tau_i,\tau_j}$ for $i \ne j$ and $i, j \in [T]$. The dissimilarity between each region $\tau_t$'s ground-truth parameter $\boldsymbol{w}_{\tau_t}^*$ and the agent's initial predictor $\boldsymbol{w}_{0}$ is upper-bounded as $||\boldsymbol{w}_{\tau_t}^*-\boldsymbol{w}_{0}||^2 \leq \Delta_{\tau_t,0}$ for $t \in [T]$.
\end{assumption}

The agent can infer each $\Delta_{\tau_i,\tau_j}$ and $\Delta_{\tau_i,0}$ for $i \ne j$, $i, j \in [T]$ based on his historical data (e.g., \cite{lin2023theory}).  
Based on our system model above, we are now ready to formulate an optimization problem for the agent in the following two stages. 
\begin{itemize}
    \item \textit{Stage I.} The agent determines a route $\boldsymbol{\tau} = (\tau_1, \cdots, \tau_T)$ over the $T$ regions to minimize an approximation of the expectation of his overall loss in \eqref{e4}.
    \item \textit{Stage II.} In each region $\tau_t$ for $t$$\in$$[T]$, he learns a predictor $\boldsymbol{w}_{\tau_t}$ to minimize an approximation of the training loss $L_{\tau_t}(\cdot)$ in \eqref{e1}.
\end{itemize}

\section{The Agent's Expected Forgetting Loss in Closed Form and NP-Hardness}\label{S3}

In this section, we first derive the closed-form formulation of the agent's expected forgetting loss in the underparameterized and overparameterized cases, respectively. Then, we prove that the optimization problem is NP-hard.  

\subsection{Analysis of the Underparameterized Case}

In Stage II, for each region $\tau_t$, $t \in [T]$, the agent aims to learn the predictor $\boldsymbol{w}_{\tau_t}$ for minimizing the training loss $L_{\tau_t}(\cdot)$ in \eqref{e1}. According to \cite{lin2023theory}, in the underparameterized case of $n \geq m +2$, minimizing \eqref{e1} returns a unique solution $\boldsymbol{w}_{\tau_t}$:
\begin{align}\label{e5}
    \boldsymbol{w}_{\tau_t} =  (\boldsymbol{x}_{\tau_t} \boldsymbol{x}_{\tau_t}^T)^{-1}\boldsymbol{x}_{\tau_t}\boldsymbol{y}_{\tau_t}.
\end{align}
After substituting $\boldsymbol{w}_{\tau_t}$ in \eqref{e5} into the agent's forgetting loss $F_T$ in \eqref{e2} and taking expectation over the feature vector $\boldsymbol{x}_{\tau_t}$ and the noise vector $\boldsymbol{z}_{\tau_t}$, we have the following.
\begin{lemma}\label{L1}
    In the underparameterized case of $n \geq m+2$, the agent's expected forgetting loss $\mathbb{E}[F_T^u]$ is in closed form:
    \begin{align}\label{e6}
\mathbb{E}[F_T^u]= 
 \sum_{i=1}^{T-1} \frac{||\boldsymbol{w}_{\tau_T}^*-\boldsymbol{w}_{\tau_i}^*||^2}{T} + \frac{m\sigma^2}{n-m-1}.
    \end{align}
\end{lemma}

Substituting $\mathbb{E}[F_T^u]$ in \eqref{e6} into \eqref{e4}, we obtain the agent's expected overall loss $\mathbb{E}[\pi_u(\boldsymbol{\tau})]$ in closed form as follows:
\begin{align}\label{e7} 
&\mathbb{E}[\pi_u(\boldsymbol{\tau})]= \nonumber \\
&  \frac{1}{T} \sum_{i=1}^{T-1} ||\boldsymbol{w}_{\tau_T}^*-\boldsymbol{w}_{\tau_i}^*||^2 + \frac{1}{T} \sum_{t=1}^{T-1}c_{\tau_t,\tau_{t+1}} + \frac{m\sigma^2}{n-m-1}.
\end{align}

Since the agent just knows each $\Delta_{\tau_i,\tau_j}$ as the upper-bound of the dissimilarity of ground-truth parameters $||\boldsymbol{w}_{\tau_i}^*-\boldsymbol{w}_{\tau_j}^*||^2$ for $i \ne j$, $i, j \in [T]$, he is only aware of an upper-bound $\mathbb{E}[\bar{\pi}_u(\boldsymbol{\tau})]$ of $\mathbb{E}[\pi_u(\boldsymbol{\tau})]$ in \eqref{e7} as below:
\begin{align}\label{e8} 
&\mathbb{E}[\bar{\pi}_u(\boldsymbol{\tau})] :=  \sum_{i=1}^{T-1} \frac{\Delta_{\tau_i,\tau_T}}{T}  + \sum_{t=1}^{T-1}\frac{c_{\tau_t,\tau_{t+1}}}{T}  + \frac{m\sigma^2}{n-m-1}. 
\end{align}
In Stage I, he focuses on minimizing the upper-bound $\mathbb{E}[\bar{\pi}_u(\boldsymbol{\tau})]$ in \eqref{e8} when determining the travelling order $\boldsymbol{\tau}$:
\begin{equation}\label{e9}
    \min_{\boldsymbol{\tau} = (\tau_1, \cdots, \tau_T)} \mathbb{E}[\bar{\pi}_u(\boldsymbol{\tau})].
\end{equation}

Note that the expected forgetting loss in \eqref{e8} only relates to the sum of task disimilarities $\Delta_{\tau_T, \tau_j}$ between the final region $\tau_T$ and any other region $\tau_j$. Thus, the optimal solution to minimize the forgetting loss is to visit region $T'$ in the end, where $T' = \arg\min_{i\in[T]}\sum_{t=1}^T\Delta_{i,t}$. However, since the agent also incurs the travelling cost, the optimization problem becomes non-trivial and the solution or algorithm is different from the traditional CL problem.

\subsection{Analysis of the Overparameterized Case}

In Stage II, for each region $\tau_t$, $t \in [T]$, the agent aims to learn the predictor $\boldsymbol{w}_{\tau_t}$ for minimizing the training loss $L_{\tau_t}(\cdot)$ in \eqref{e1}. According to \cite{lin2023theory}, in the overparameterized case of $m \geq n + 2$,
minimizing \eqref{e1} returns infinite solutions with zero loss. To preserve as much information about old tasks as possible, here we focus on the solution that has the smallest $\ell_2$-norm distance with $\boldsymbol{w}_{\tau_{t-1}}$, which is the convergent point of the stochastic gradient descent method as follows:
\begin{align}\label{e11}
    \boldsymbol{w}_{\tau_t} = \boldsymbol{w}_{\tau_{t-1}} + \boldsymbol{x}_{\tau_t} (\boldsymbol{x}_{\tau_t}^T \boldsymbol{x}_{\tau_t})^{-1} (\boldsymbol{y}_{\tau_t} - \boldsymbol{x}_{\tau_t}^T \boldsymbol{w}_{\tau_{t-1}}).
\end{align}

Define $r$$:=$$1$$-$$\frac{n}{m}$. After substituting $\boldsymbol{w}_{\tau_t}$ in \eqref{e11} into forgetting loss $F_T$ in \eqref{e2} and taking expectation over the feature vector $\boldsymbol{x}_{\tau_t}$ and the noise vector $\boldsymbol{z}_{\tau_t}$, we have the following.
\begin{lemma}\label{L2}
    In the overparameterized case of $m \geq n+2$, the agent's expected forgetting loss $\mathbb{E}[F_T^o]$ is in closed form:
    \begin{align}
\mathbb{E}[F_T^o]=& 
 \sum_{i=1}^T\frac{(1-r)r^{T-i}}{T}\sum_{j=1}^T ||\boldsymbol{w}_{\tau_i}^*-\boldsymbol{w}_{\tau_j}^*||^2 \nonumber \\
 &+\frac{r^T}{T} \sum_{i=1}^T ||\boldsymbol{w}_{\tau_i}^*-\boldsymbol{w}_0||^2  +  \frac{(1-r^T)m\sigma^2}{m-n-1}.\label{e12}
    \end{align}
\end{lemma}

Substituting $\mathbb{E}[F_T^o]$ in \eqref{e12} into \eqref{e4} and replace each $||\boldsymbol{w}_{\tau_i}^*-\boldsymbol{w}_{\tau_j}^*||^2$ with the agent's known upper-bound $\Delta_{\tau_i,\tau_j}$, we obtain an upper-bound $\mathbb{E}[\bar{\pi}_o(\boldsymbol{\tau})]$ of the agent's expected overall loss:
\begin{align} 
\mathbb{E}[\bar{\pi}_o(\boldsymbol{\tau})]
 =& \sum_{i=1}^T\frac{(1-r)r^{T-i}}{T}\sum_{j=1}^T \Delta_{\tau_i,\tau_j} + \sum_{t=1}^{T-1}\frac{c_{\tau_t,\tau_{t+1}}}{T} \nonumber \\
    &+\frac{r^T}{T} \sum_{i=1}^T \Delta_{\tau_i,0}  +  \frac{(1-r^T)m\sigma^2}{m-n-1}. \label{e14}
\end{align}
Since the agent incurs the travelling cost, the optimization problem to minimize \eqref{e14} becomes non-trivial and the solution or algorithm is different from the traditional CL problem.

\subsection{NP-Hardness}

One may wonder if we can efficiently solve the optimization problem of minimizing \eqref{e8} or \eqref{e14} in a polynomial time. Since objective functions $\mathbb{E}[\bar{\pi}_u(\boldsymbol{\tau})]$ in \eqref{e8} and $\mathbb{E}[\bar{\pi}_o(\boldsymbol{\tau})]$ in \eqref{e14} both contain the term $\sum_{t=1}^{T-1}\frac{c_{\tau_t,\tau_{t+1}}}{T}$, to solve either problem is at least as hard as solving the following problem:
\begin{equation}\label{e10}
    \min_{\boldsymbol{\tau} = (\tau_1, \cdots, \tau_T)} \sum_{t=1}^{T-1}\frac{c_{\tau_t,\tau_{t+1}}}{T}.
\end{equation}
Note that the problem in \eqref{e10} is same as the classic shortest Hamiltonian path (SHP) problem in a graph $G = (V, E)$, where we denote each region $\tau_t$ as a vertex and denote each $c_{\tau_i,\tau_j}$ as the weight of edge $e_{\tau_i,\tau_j}$. Since the SHP problem is known as NP-hard (e.g., \cite{sevaux2008hamiltonian}), we have the following.
\begin{proposition}\label{prop1}
The agent's problem of minimizing either \eqref{e8} in the underparameterized case or \eqref{e14} in the overparameterized case is NP-hard.
\end{proposition}

Proposition~\ref{prop1} indicates that we cannot find the optimal solution to our problem in a polynomial time. Therefore, we are motivated to propose an efficient algorithm to find an approximation solution within a polynomial time in the region number $T$, guaranteed with a small approximation ratio.

\section{An Efficient Approximation Algorithm Design, Analysis and Simulation}\label{S4}

In this section, we first present our algorithm design and its approximation ratios in the underparameterized and overparameterized cases, respectively. We then run simulations to verify our algorithm's close-to-optimum approximation.

\subsection{Algorithm Design for the Underparameterized Case}
We first present our algorithm design for the underparameterized case. According to the objective function $\mathbb{E}[\bar{\pi}_u(\boldsymbol{\tau})]$ in \eqref{e8}, we find that only the first and the second terms of $\mathbb{E}[\bar{\pi}_u(\boldsymbol{\tau})]$ depend on the travelling order $\boldsymbol{\tau}$. Further, the optimal solution to minimize the first term $\sum_{i=1}^{T-1} \frac{\Delta_{\tau_i,\tau_T}}{T}$ is to visit region $T'$ in the end, where $T' = \arg\min_{i\in[T]}\sum_{t=1}^T\Delta_{i,t}$. Minimizing the second term is the same as the NP-hard SHP in \eqref{e10}, where we can introduce a variance of the Christofides' Algorithm (e.g., \cite{van2020historical}) to return a $\frac{3}{2}$-approximated solution. 

Based on the above analysis, we propose Algorithm~\ref{Al1} as an approximated algorithm to efficiently solve the problem in \eqref{e9}, which contains a two-layer design: first, we use Algorithm~\ref{Al1} to obtain a  $\frac{3}{2}$-approximated solution to the problem in \eqref{e10}; further, we make sure that the output solution of  Algorithm~\ref{Al1} ends with region $T'$ to minimize the forgetting-loss part in \eqref{e8}. 

\begin{algorithm}
\caption{An Approximation Algorithm}
\label{Al1}
\begin{algorithmic}[1]
\REQUIRE Graph $G = (V, E)$, where the vertex number $|V|=T$, each edge's weight $e_{j,k} = c_{j,k}$ for all $j,k \in V$. Graph $G' = (V', E')$, where the vertex number $|V'|=T+1$, each edge's weight $e_{j,k} = c_{j,k}$ for all $j,k \in V'$, $j \ne v_0$, $k \ne v_0$, and $e_{v, v_0}=0$ for any $v \in V$ and $v \ne v_0$;
\ENSURE A route $(\tau_1, \cdots, \tau_{T})$.
\STATE Create a minimum spanning tree $T$ of $G$ and add the vertex $v_0$ to vertex $v'$ to form a new graph $T'$, where $v' = \arg\min_{i\in[T]} \sum_{t=1}^T \Delta_{i, t}$;
\STATE Find a minimum-weight perfect matching $M$ in the subgraph induced in $G'$ by $O$, where $O$ is the set of vertices with odd degree in $T'$;
\STATE Combine the edges of $M$ and $T'$ to form a connected multigraph $H$ in which each vertex has even degree;
\STATE Form an Eulerian circuit in $H$;
\STATE Make the circuit found in previous step into a Hamiltonian circuit by skipping repeated vertices not connected with $v_0$ (shortcutting) and keeping the edge $e_{v', v_0}$;
\STATE Remove the edges connected to the vertex $v_0$ to obtain a Hamiltonian path ending with vertex $v'$.
\RETURN The obtained Hamiltonian path.
\end{algorithmic}
\end{algorithm}

\begin{proposition}\label{prop3}
In the underparameterized case of $n \geq m+2$, the output of our Algorithm~\ref{Al1} incurs an approximation ratio of $\frac{3}{2}$ to the optimum of the agent's expected overall loss in \eqref{e8}, with a complexity order of $\mathcal{O}(T^3)$ in region number $T$.
\end{proposition}

\subsection{Algorithm Design for the Overparameterized Case}

We then present our algorithm design for the overparameterized case. According to the objective function $\mathbb{E}[\bar{\pi}_o(\boldsymbol{\tau})]$ in \eqref{e14}, we find that only the first and the second terms of $\mathbb{E}[\bar{\pi}_o(\boldsymbol{\tau})]$ depend on the travelling order $\boldsymbol{\tau}$. Further, the optimal solution to minimize the first term $\sum_{i=1}^T\frac{(1-r)r^{T-i}}{T}\sum_{j=1}^T \Delta_{\tau_i,\tau_j}$ is to route in an order of 
\begin{align}\label{e15}
    \boldsymbol{\tau}' := (1', \cdots, T'),
\end{align}
where $\sum_{t=1}^T\Delta_{1', t} \geq \cdots \geq \sum_{t=1}^T\Delta_{T', t}$. Minimizing the second term is the same as the NP-hard SHP in \eqref{e10}, where we can use Algorithm~\ref{Al1} to return a $\frac{3}{2}$-approximated solution. 

Though our Algorithm~\ref{Al1} may not output the exact same order as  $\boldsymbol{\tau}'$ in \eqref{e15} for minimizing the forgetting loss, it can at least guarantee the last region same as that of $\boldsymbol{\tau}'$. 

\begin{proposition}\label{prop4}
In the overparameterized case of $m \geq n+2$, the output of our Algorithm~\ref{Al1} incurs an approximation ratio of $\frac{3}{2}+r^{1-T}$ to the optimum of the agent's expected overall loss in \eqref{e8}, with a complexity order of $\mathcal{O}(T^3)$ in region number $T$. Besides, if $m \gg n$, i.e., the feature number $m$ is sufficiently larger than the sample number $n$, the ratio improves to $\frac{3}{2}$. 
\end{proposition}

If the feature number $m$ is sufficiently larger than the sample number $n$, we have $r$$=$$1-n/m$$\to$$1$. The first term $\sum_{i=1}^T\frac{(1-r)r^{T-i}}{T}\sum_{j=1}^T \Delta_{\tau_i,\tau_j}$ in \eqref{e14} 
related to forgetting loss tends to zero, indicating that more features alleviate the negative impact of region dissimilarity on the forgetting loss. Therefore, the problem degenerates to \eqref{e10} and our Algorithm~\ref{Al1} achieves a smaller approximation ratio of $\frac{3}{2}$.

According to Propositions~\ref{prop3} and \ref{prop4}, we have the following.
\begin{corollary}
  If the region number $T$=2, our Algorithm~\ref{Al1} always returns the optimal solution to the optimization problem in both the underparameterized and overparameterized cases.
\end{corollary}

If the region number $T$=2, we can find that the optimal solution to minimize both $\mathbb{E}[\bar{\pi}_u(\boldsymbol{\tau})]$ in \eqref{e8} and $\mathbb{E}[\bar{\pi}_o(\boldsymbol{\tau})]$ in \eqref{e14} is to visit region $T' = \arg\min_{i\in[T]}\sum_{t=1}^T\Delta_{i,t}$ in the end, which is the same as the output solution of our Algorithm~\ref{Al1}. It then achieves the optimum.

\subsection{Simulation Results for Verification}

In this section, we run simulations to show our algorithm~\ref{Al1}'s even smaller ratios than the theoretical bounds. We randomly generate each upper-bound $\Delta_{\tau_i, \tau_j}$, $\Delta_{\tau_i, 0}$ and each travelling cost $c_{\tau_i,\tau_j}$ in the range of [1,10] without loss of generality. We fix the sample number $n=100$, change the feature number $m$ and the region number $T$ to verify for the underparameterized and the overparameterized cases, respectively. We define $R:= \mathbb{E}[\bar{\pi}^*]/\mathbb{E}[\bar{\pi}^{**}]$ as the ratio of the agent's expected overall loss  $\mathbb{E}[\bar{\pi}^*]$ of our Algorithm~\ref{Al1} to the optimum $\mathbb{E}[\bar{\pi}^{**}]$, where $R$$\geq$1.

\begin{figure}
    \centering
    \includegraphics[width=0.42\linewidth]{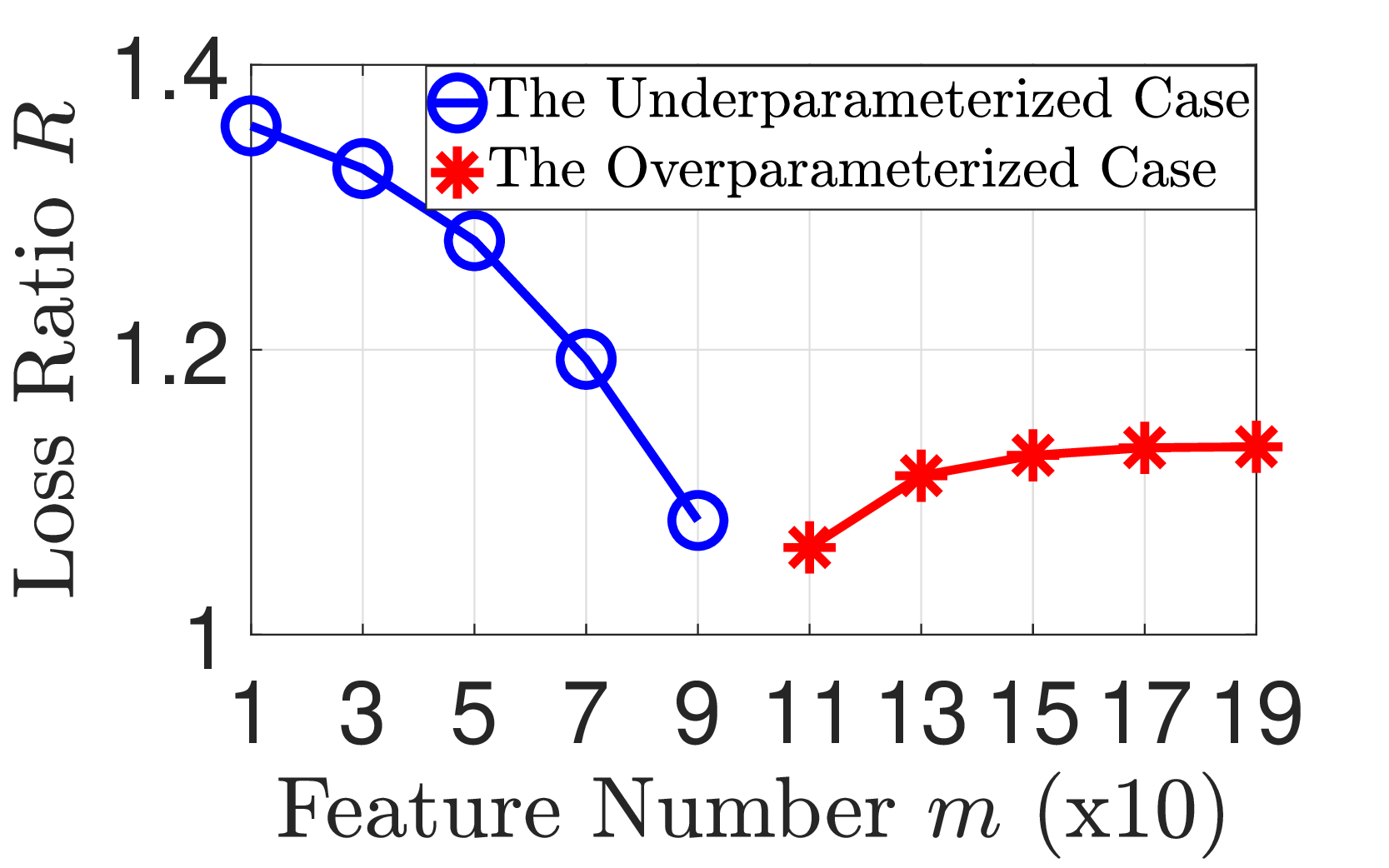}
    \caption{The ratio $R$ between the agent's expected overall loss $\mathbb{E}[\bar{\pi}^*]$ of our Algorithm~\ref{Al1} and the optimum $\mathbb{E}[\bar{\pi}^{**}]$ versus the feature number $m$. }
    \label{fig1}
\end{figure}

\begin{figure}
    \centering
    \includegraphics[width=0.42\linewidth]{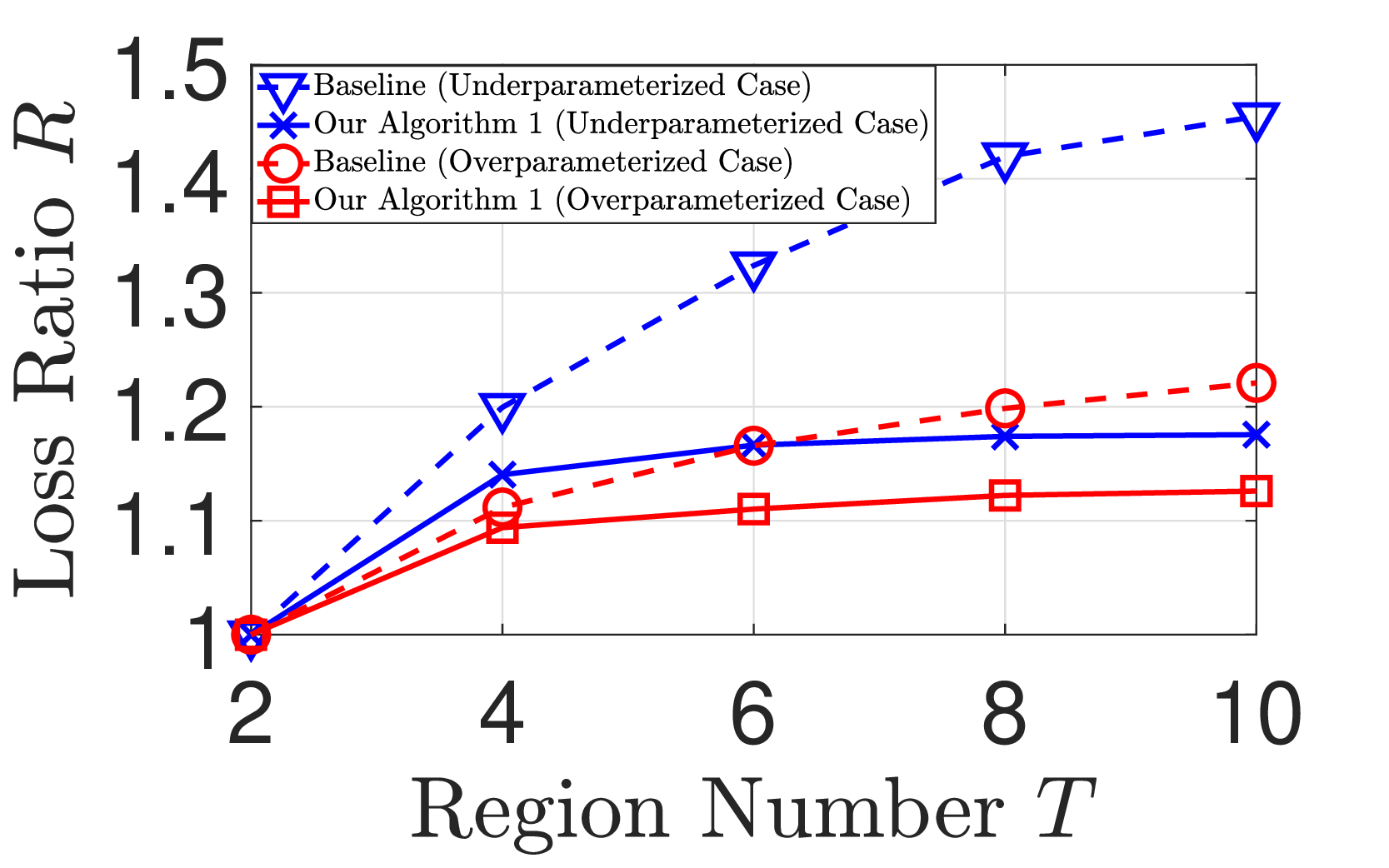}
    \caption{The ratios between the agent's expected overall loss $\mathbb{E}[\bar{\pi}^*]$ of our Algorithm~\ref{Al1} and the CL baseline to the optimum $\mathbb{E}[\bar{\pi}^{**}]$ versus the region number $T$, respectively. Here we choose feature number $m$=80 for the underparameterized case and $m$=120 for the overparameterized case.}
    \label{fig2}
\end{figure}

Figure~\ref{fig1} shows the loss ratio $R$ versus the feature number $m$. Since our Algorithm~\ref{Al1} mainly focuses on approximating the optimal order for minimizing the travelling cost, the loss ratio $R$ may not be monotonic in $m$. Nevertheless, we find that $R$ in Figure~\ref{fig1} is still smaller than $\frac{3}{2}$ of Propositions~\ref{prop3} and \ref{prop4}, which implies our Algorithm~\ref{Al1}'s good approximation.

Figure~\ref{fig2} shows the loss ratio $R$ versus the region number $T$ in the underparameterized and overparameterized cases under our Algorithm~\ref{Al1} and the CL baseline, respectively, in which the agent aims to minimize the expected forgetting loss (e.g., \cite{evron2022catastrophic}, \cite{lin2023theory}). It demonstrates our algorithm's over 50\% improvement in the underparameterized case and over 30\% improvement in the overparameterized case for $T$$\geq$6,  respectively. Further, we find that our Algorithm~\ref{Al1}'s approximation ratio is smaller than $\frac{3}{2}$ of Propositions~\ref{prop3} and \ref{prop4}, implyng its good  approximation.

\section{Conclusion}
In this paper, we study how to opportunistically route the testing object and alter the task sequence in CL. We formulate it as a new optimization problem and prove it NP-hard. We propose an algorithm to achieve certain approximation ratios in a polynomial time in task number $T$.
Simulation results verify our algorithm’s close-to-optimum performance. 

\newpage


\newpage
\onecolumn
\appendix 

\subsection{Proof of Lemma~\ref{L1}}

Please refer to Appendix D.8 in reference \cite{lin2023theory}.

\subsection{Proof of Lemma~\ref{L2}}

Please refer to Appendix D.3 in reference \cite{lin2023theory}.

\subsection{Proof of Proposition~\ref{prop3}}

Before we formally prove Proposition~\ref{prop3}, let us introduce a useful lemma with a formal proof first. 

\begin{lemma}\label{L3}
   The output of Algorithm~\ref{Al1} incurs an approximation ratio of $\frac{3}{2}$ to the optimum of the problem in \eqref{e10}. 
\end{lemma}
\textit{Proof.} First, we have the weight $W(T')$ of the graph $T'$ is same as that $W(T)$ of the minimum spanning tree (MST) $T$ due to $e_{v, v_0} = 0$. Denote OPT as the weight of the optimal Hamiltonian path or solution to the SHP in \eqref{e10}, we then have
\begin{align*}
    W(T') = W(T) \leq \text{OPT}
\end{align*}
since the weight of a Hamiltonian path (as a spanning tree) is no larger than that of a MST. 

Further, according to \cite{van2020historical}, the weight $W(M)$ of a minimum-weight perfect matching $M$ in the subgraph induced in $G'$ by $O$ is no larger than half of that of the optimal Hamiltonian cycle of the graph $G'$. Since $e_{v, v_0}=0$ for all $v \in V'$ and $v \ne v_0$ in the graph $G'$, we have the weight of the optimal Hamiltonian cycle of the graph $G'$ is the same as that of the optimal Hamiltonian path of the graph $G$, which implies
\begin{align*}
    W(M) \leq \frac{1}{2}\text{OPT}.
\end{align*}
After shortcutting the multigraph $H$ (obtained by combining the edges of $M$ and $T'$) to obtain a Hamiltonian circuit $C$, we have the weight of $C$ can only be decreased due to the triangle inequality $c_{ij} \leq c_{ik} + c_{kj}$ for $i,j,k \in V$ and $i \ne j \ne k$, i.e., 
\begin{align*}
    W(C) \leq W(T') + W(M)\leq \frac{3}{2}\text{OPT}.
\end{align*}

After removing the edges connected to the vertex $v_0$ in the Hamiltonian circuit $C$  to obtain a
Hamiltonian path $P$ ending with vertex $v'$, we have the weight of $P$ is the same as that of $C$ due to $e_{v, v_0}=0$ for all $v \in V'$ and $v \ne v_0$. Finally, we have
\begin{align*}
    W(P) = W(C) \leq \frac{3}{2}\text{OPT}.
\end{align*}
We then finish the proof. $\hfill\square$

We then formally prove Proposition~\ref{prop3}.  Denote $\boldsymbol{\tau^*}$ as the route returned by Algorithm~\ref{Al1} and $\boldsymbol{\tau^{**}}$ as the optimal route. We have
    \begin{align*}
        \frac{\mathbb{E}[\bar{\pi}_u(\boldsymbol{\tau^*})]}{\mathbb{E}[\bar{\pi}_u(\boldsymbol{\tau^{**}})]}= \frac{\mathbb{E}[F_T(\boldsymbol{w}_{\tau_T^*})] + C_T(\boldsymbol{\tau^*})}{\mathbb{E}[F_T(\boldsymbol{w}_{\tau_T^{**}})] + C_T(\boldsymbol{\tau^{**}})} < \frac{C_T(\boldsymbol{\tau^*})}{C_T(\boldsymbol{\tau^{**}})} \leq \frac{3}{2}.
    \end{align*}
    The first inequality holds because $\mathbb{E}[F_T(\boldsymbol{w}_{\tau_T^{*}})]$ is minimized by ending with region $T'$ and thus $\mathbb{E}[F_T(\boldsymbol{w}_{\tau_T^{*}})] \leq \mathbb{E}[F_T(\boldsymbol{w}_{\tau_T^{**}})]$. The second inequality holds due to Lemma~\ref{L3}. Therefore, we have Algorithm~\ref{Al1} returns a solution of an approximation ratio 
 $\frac{3}{2}$ to the optimum. The complexity order $\mathcal{O}(T^3)$ comes from finding the minimum perfect matching $M$, which has been proved in \cite{van2020historical}. We then finish the proof.

\subsection{Proof of Proposition~\ref{prop4}} 

We first prove the ratio in the general overparameterized case $m \geq n + 2$. Denote $\boldsymbol{\tau^*}$ as the training order returned by Algorithm~\ref{Al1} and $\boldsymbol{\tau^{**}}$ as the optimal training order. We have
    \begin{align*}
        \frac{\mathbb{E}[\bar{\pi}(\boldsymbol{\tau^*})]}{\mathbb{E}[\bar{\pi}(\boldsymbol{\tau^{**}})]}&= \frac{\mathbb{E}[F_T(\boldsymbol{w}_{\tau_T^*})] + \mathbb{E}[C_T(\boldsymbol{\tau^*})]}{\mathbb{E}[F_T(\boldsymbol{w}_{\tau_T^{**}})] + \mathbb{E}[C_T(\boldsymbol{\tau^{**}})]} \\
        &< \frac{\frac{1}{T}\sum_{i=1}^T(1-r)r^{T-i}\sum_{j=1}^T \Delta_{\tau_i^*, \tau_j^*}+\mathbb{E}[C_T(\boldsymbol{\tau^*})]}{\frac{1}{T}\sum_{i=1}^T(1-r)r^{T-i}\sum_{j=1}^T \Delta_{\tau_i^{**}, \tau_j^{**}}+\mathbb{E}[C_T(\boldsymbol{\tau^{**}})]} \\
        &< \frac{\frac{1}{T}\sum_{i=1}^T(1-r)\sum_{j=1}^T \Delta_{\tau_i^*, \tau_j^*}+\mathbb{E}[C_T(\boldsymbol{\tau^*})]}{\frac{1}{T}\sum_{i=1}^T(1-r)r^{T-1}\sum_{j=1}^T \Delta_{\tau_i^{**}, \tau_j^{**}}+\mathbb{E}[C_T(\boldsymbol{\tau^{**}})]} \\
        &< \frac{\frac{1}{T}\sum_{i=1}^T(1-r)\sum_{j=1}^T \Delta_{\tau_i^*, \tau_j^*}}{\frac{1}{T}\sum_{i=1}^T(1-r)r^{T-1}\sum_{j=1}^T \Delta_{\tau_i^{**}, \tau_j^{**}}} + \frac{\mathbb{E}[C_T(\boldsymbol{\tau^*})]}{\mathbb{E}[C_T(\boldsymbol{\tau^{**}})]} \\
        &= r^{1-T} + \frac{\mathbb{E}[C_T(\boldsymbol{\tau^*})]}{\mathbb{E}[C_T(\boldsymbol{\tau^{**}})]} \leq r^{1-T} + \frac{3}{2}.
    \end{align*}
    The first inequality holds because we subtract the common term $\frac{r^T}{T} \sum_{i=1}^T \Delta_{\tau_i, 0}+ \frac{(1-r^T)m\sigma^2}{m-n-1}$ which is independent of the order. The second inequality holds due to $r^{T-i} \in [r^{T-1}, 1]$. The third inequality holds due to $\frac{a+b}{c+d} < \frac{a}{c}+\frac{b}{d}$ if $a,b,c,d>0$. The second equality holds since the term $\frac{1}{T}\sum_{i=1}^T\sum_{j=1}^T \Delta_{\tau_i, \tau_j}$ has the same value regardless of training order. The last inequality holds due to Lemma~\ref{L3}. Therefore, we have Algorithm~\ref{Al1} returns a solution with $o(\frac{3}{2}+r^{1-T})$ approximation ratio to the optimum. The complexity order $\mathcal{O}(T^3)$ comes from finding the minimum perfect matching $M$, which has been proved in \cite{van2020historical}.

If $m \gg n$, we have $r=1-n/m \to 1$ and the objective function $\mathbb{E}[\bar{\pi}_o(\boldsymbol{\tau})]$ in \eqref{e14} becomes
\begin{align} 
\mathbb{E}[\bar{\pi}_o(\boldsymbol{\tau})]
 = \sum_{t=1}^{T-1}\frac{c_{\tau_t,\tau_{t+1}}}{T} +\frac{1}{T} \sum_{i=1}^T \Delta_{\tau_i,0}. \nonumber
\end{align}
Denote $\boldsymbol{\tau^*}$ as the route returned by Algorithm~\ref{Al1} and $\boldsymbol{\tau^{**}}$ as the optimal route. We have
    \begin{align*}
        \frac{\mathbb{E}[\bar{\pi}_o(\boldsymbol{\tau^*})]}{\mathbb{E}[\bar{\pi}_o(\boldsymbol{\tau^{**}})]}= \frac{\frac{1}{T} \sum_{i=1}^T \Delta_{\tau_i^*,0} + C_T(\boldsymbol{\tau^*})}{\frac{1}{T} \sum_{i=1}^T \Delta_{\tau_i^{**},0} + C_T(\boldsymbol{\tau^{**}})} < \frac{C_T(\boldsymbol{\tau^*})}{ C_T(\boldsymbol{\tau^{**}})} \leq \frac{3}{2}.
    \end{align*}
    The first inequality holds because we subtract the common term $\frac{1}{T} \sum_{i=1}^T \Delta_{\tau_i,0}$ which is independent of the travelling order. The second inequality holds due to Lemma~\ref{L3}. Therefore, we have Algorithm~\ref{Al1} returns a solution of an approximation ratio $\frac{3}{2}$ to the optimum.  We then finish the proof.

\end{document}